\documentclass[journal]{IEEEtran}
\usepackage[hidelinks]{hyperref}
\usepackage{orcidlink}
\usepackage{graphicx} 
\usepackage[shortcuts,acronym]{glossaries}
\usepackage{booktabs} 
\usepackage{multirow}
\usepackage{tikz}
\usepackage{cleveref}
\usepackage[numbers,sort&compress]{natbib}
\usepackage{url}
\usepackage{xcolor}

\definecolor{DarkGreen}{RGB}{0,150,0}
\definecolor{DarkRed}{RGB}{180,0,0}
\definecolor{DarkGray}{gray}{0.5}

\Crefname{figure}{Fig.}{Figs.}

\newacronym{ads}{ADS}{Automated driving system}
\newacronym{bev}{BEV}{bird’s-eye view}
\newacronym{av}{AV}{automated vehicle}
\newacronym{hd}{HD}{high-definition}
\newacronym{td}{top-down}{top-down}
\newacronym{map}{mAP}{mean Average Precision}
\newacronym{iou}{IoU}{Intersection over Union}
\newacronym{uav}{UAV}{unmanned aerial vehicle}
\newacronym{gsd}{GSD}{ground sample distance}
\newacronym{rtk}{RTK}{Real-Time Kinematic}
\newacronym{ppk}{PPK}{Post-Processing Kinematic}
\newacronym{minade}{minADE}{minimum Average Displacement Error}
\newacronym{minfde}{minFDE}{minimum Final Displacement Error}
\newacronym{mr}{MR}{Miss Rate}

\usepackage{eso-pic}
\usepackage{graphicx}

\newcommand\PreprintHeader{%
  \put(0,760){%
    \parbox{\paperwidth}{%
      \centering
      \small\itshape
      This work has been submitted to the IEEE for possible publication.\\
      Copyright may be transferred without notice, after which this version may no longer be accessible.
    }%
  }%
}

\AddToShipoutPictureBG*{%
  \PreprintHeader
}

\begin{document}
\bstctlcite{main:BSTcontrol}

\title{AID4AD: Aerial Image Data for Automated Driving Perception}

\author{Daniel~Lengerer~\orcidlink{0009-0009-1600-6144},
        Mathias Pechinger~\orcidlink{0000-0003-2371-9870},
        Klaus Bogenberger~\orcidlink{0000-0003-3868-9571},
        Carsten Markgraf~\orcidlink{0000-0001-9447-2065}
\thanks{This work was supported by the Hightech Agenda Bavaria, funded by the Free State of Bavaria, Germany.}
\thanks{D. Lengerer and C. Markgraf are with the Faculty of Electrical Engineering,
        Technical University of Applied Sciences Augsburg, Germany
            {\tt\small  \{Daniel.Lengerer, Carsten.Markgraf\}@tha.de}}%
\thanks{M. Pechinger and K. Bogenberger are with the Chair of Traffic Engineering and Control,
        Technical University of Munich, Germany
            {\tt\small \{Mathias.Pechinger, Klaus.Bogenberger\}@tum.de}}%
}


\maketitle
\begin{abstract}
This work investigates the integration of spatially aligned aerial imagery into perception tasks for \acp{av}. As a central contribution, we present AID4AD, a publicly available dataset that augments the nuScenes dataset with high-resolution aerial imagery precisely aligned to its local coordinate system. The alignment is performed using SLAM-based point cloud maps provided by nuScenes, establishing a direct link between aerial data and nuScenes local coordinate system.
To ensure spatial fidelity, we propose an alignment workflow that corrects for localization and projection distortions. A manual quality control process further refines the dataset by identifying a set of high-quality alignments, which we publish as ground truth to support future research on automated registration.
We demonstrate the practical value of AID4AD in two representative tasks: in online map construction, aerial imagery serves as a complementary input that improves the mapping process; in motion prediction, it functions as a structured environmental representation that replaces \acl{hd} maps. Experiments show that aerial imagery leads to a 15–23\% improvement in map construction accuracy and a 2\% gain in trajectory prediction performance.
These results highlight the potential of aerial imagery as a scalable and adaptable source of environmental context in \acl{av} systems, particularly in scenarios where \acl{hd} maps are unavailable, outdated, or costly to maintain. AID4AD, along with evaluation code and pretrained models, is publicly released to foster further research in this direction: \url{https://github.com/DriverlessMobility/AID4AD}.

\end{abstract}

\section{Introduction}
\label{sec:intro}

\IEEEPARstart{A}{erial} imagery has seen major advances in recent years, driven by improvements in spatial resolution, georeferencing accuracy, and platform diversity. Modern satellites such as WorldView-3 now deliver \acp{gsd} of 25--30\,cm, exceeding the 50\,cm standard of a decade ago and enabling the identification of small-scale features from orbit~\cite{WorldView3_ESA}. Likewise, photogrammetric pipelines using crewed aircraft or \acp{uav} can achieve sub-10\,cm resolution and centimeter-level positioning when enhanced with \ac{rtk} or \ac{ppk} GNSS corrections~\cite{Jasnoch_2023}.

These capabilities support diverse applications. In precision agriculture, \ac{uav} imagery enables early crop stress detection and targeted treatment~\cite{PrecisionAg_Review}. In environmental monitoring, high-resolution satellite imagery is used to assess deforestation, urban growth, and vegetation dynamics~\cite{RemoteSensing_Applications}. 
In traffic behavior research, recent large-scale \ac{uav}-based datasets enable the continuous observation of complex urban traffic scenes, capturing interactions between motorized vehicles, cyclists, and pedestrians with high spatiotemporal fidelity \cite{kutsch_tumdotmuc_2024}.

Public agencies also adopted these technologies. National datasets such as Germany’s DOP20, the UK’s OS MasterMap Imagery Layer, and France’s BD ORTHO provide regularly updated aerial imagery at 20--25\,cm resolution, and are widely used in urban, cadastral, and infrastructure planning~\cite{DOP20_BKG, OS_UK, BDORTHO_FR}.

These developments underscore the potential of aerial imagery as a reliable, high-resolution, and geospatially accurate data source. This work investigates its suitability for supporting localization, perception, and scene understanding in \ac{av} systems, and assesses its feasibility as a scalable environmental reference for automated driving.

\Acp{av} rely on detailed understanding of the static environment, typically provided by \ac{hd} maps. These maps support motion planning and also enhance perception tasks such as motion prediction by providing structured lane and intersection context~\cite{multipath, multipathpp, zhou2022hivt}.

\begin{figure}[t]
    \centering
    \begin{minipage}[b]{0.23\textwidth}
        \centering
        \includegraphics[width=\textwidth]{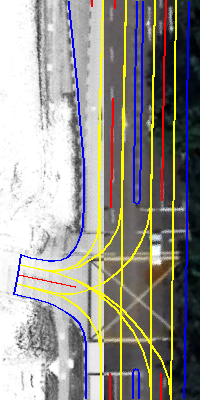}
    \end{minipage}
    \hfill
    \begin{minipage}[b]{0.23\textwidth}
        \centering
        \includegraphics[width=\textwidth]{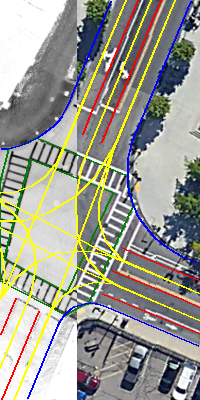}
    \end{minipage}
    \caption{
    Examples from the proposed dataset, showing map annotations overlaid on the base map and the aligned aerial image. Aerial imagery: Google Earth, Image © 2024 Maxar Technologies.}
    \label{fig:alignment}
\end{figure}

While \ac{hd} maps enable state-of-the-art performance, their maintenance poses significant challenges, especially in dynamic urban environments. As discussed in~\cite{hdmapchallenges}, \ac{hd} maps require specialized mapping vehicles, expert validation, and extensive post-processing, making them costly and difficult to scale. Moreover, incorporating rapid road changes is complex due to the time-intensive data collection and update process. Although dynamic updates have been explored, they remain constrained by limited data frequency and processing delays.

We explore whether aerial imagery can serve as a complementary or alternative source of environmental context. Its top-down perspective offers broad coverage and, particularly through drones, the potential for near real-time updates to reflect temporary changes such as roadworks or accidents. Satellite and aircraft imagery can also support perception in areas where \ac{hd} maps are unavailable or incomplete.

However, it remains unclear whether aerial imagery provides sufficient detail for \ac{av} perception. A proof-of-concept is needed to evaluate its ability to support tasks such as motion prediction and scene understanding before addressing challenges like precise localization or dynamic map updates.

Existing datasets that combine aerial imagery and \ac{av} sensor data often lack precise spatial alignment and exhibit significant temporal mismatches. These limitations lead to inconsistencies in how the road network is represented from above, making it difficult to assess the feasibility of aerial imagery for perception tasks. To the best of our knowledge, no publicly available dataset offers spatially aligned aerial imagery suitable for investigating its use as a substitute for \ac{hd} maps.

To address this gap, we collect high-resolution aerial imagery, up to 0.15\,m per pixel, from satellite and aircraft sources and align it precisely with the nuScenes dataset~\cite{nuscenes}. Imagery is sourced from Google Earth Pro, based on Google Earth Engine~\cite{googleearthengine}, and selected to closely match the nuScenes recording period. We also present a methodology for spatially aligning aerial and vehicle data, enabling an initial feasibility assessment prior to constructing a dedicated joint dataset.

The main contributions of this work are:

\begin{enumerate}
\item We demonstrate that high-resolution aerial imagery can serve as a structured and scalable environmental reference for automated driving perception.
\item We introduce and publicly release AID4AD, the first dataset of spatially aligned aerial images integrated with the nuScenes benchmark.
\item We propose a workflow for registering aerial imagery to the local coordinate system of SLAM-based point cloud maps, enabling precise alignment.
\item We evaluate the use of aerial imagery in two key perception tasks:
\begin{enumerate}
\item \textbf{Online Map Construction}: augmenting map generation using aerial data.
\item \textbf{Motion Prediction}: replacing \acl{hd} maps as environmental information source.
\end{enumerate}
\end{enumerate}

This study bridges the gap between \ac{hd} map-based and aerial imagery-based perception by demonstrating feasibility and providing an open dataset to foster further research.

The remainder of this article is structured as follows. \Cref{sec:related_work} reviews related work, including datasets, online map construction, and trajectory prediction algorithms. \Cref{sec:AID4AD_collection} details the creation of the AID4AD dataset, covering the motivation for alignment, temporal considerations, the alignment workflow, and evaluation. \Cref{sec:meth_algos_eval} presents the methodology used to evaluate aerial imagery within state-of-the-art algorithms. Experimental results and key findings are reported in \cref{sec:results}. Finally, \Cref{sec:conclusion} concludes the paper and outlines directions for future research.

\section{Related Work}
\label{sec:related_work}

\subsection{Datasets}

Aerial imagery has been widely used across various research domains, particularly in road network extraction and map construction. Prior works \cite{aid_map_constr, aid_map_constr2} have demonstrated the feasibility of deriving lane geometries, intersections, and road connectivity from aerial images using deep learning techniques. These methods highlight the potential of aerial imagery to serve as a structured environmental representation, similar to \ac{hd} maps, without requiring extensive ground-based surveying.

Despite this potential, few efforts have attempted to integrate aerial imagery with \ac{av} datasets. Two notable works, SatforHDMap \cite{satforhd} and OpenSatMap \cite{opensatmap}, have sought to incorporate satellite imagery into \ac{av} research. SatforHDMap provides aerial image crops for each keyframe in nuScenes, facilitating their use in map construction algorithms such as HDMapNet \cite{li2021hdmapnet}. OpenSatMap, on the other hand, is not explicitly linked to any \ac{av} dataset but offers annotated satellite images spanning 60 cities and 19 countries, including locations covered by nuScenes and Argoverse. To demonstrate its applicability in \ac{av} research, OpenSatMap applies the same adapted HDMapNet model as SatforHDMap for automated map generation.

However, both datasets present significant limitations that restrict their ability to evaluate aerial imagery as an alternative to \ac{hd} maps. Their spatial misalignment with \ac{av} sensor data and temporal discrepancies between aerial image acquisition and \ac{av} dataset recordings introduce inconsistencies in road network representation. These limitations make it difficult to systematically assess aerial imagery's feasibility in perception tasks.

To address these challenges, we introduce the AID4AD dataset, ensuring precise spatial alignment of aerial images with \ac{av} sensor data while selecting images recorded as close in time as possible to the nuScenes dataset.

For integrating aerial imagery with \ac{av} datasets, we considered several large-scale \ac{av} datasets, each with different trade-offs in sensor coverage, dataset size, and localization accuracy. The most widely adopted datasets include nuScenes \cite{nuscenes}, Waymo Open Dataset \cite{waymo_perc}, and KITTI \cite{Geiger2013IJRR}.

Waymo Open Dataset provides a larger-scale dataset with a similar sensor setup to nuScenes, but it does not provide GNSS coordinates, making precise aerial image alignment difficult.
KITTI, in contrast, offers accurate localization information. Still, its limited sensor coverage—notably the lack of surround-view cameras—makes it less suitable for perception tasks that rely on a complete environmental understanding.
Given these trade-offs, we selected nuScenes as the basis for our work. It offers the best balance between dataset size, sensor coverage, and localization accuracy. nuScenes provides multi-sensor data (LiDAR, cameras, radar) and includes geographically well-defined recording locations, making it the most suitable choice for integrating aerial imagery.

\subsection{Online Map Construction}

Online map construction algorithms aim to generate vectorized environmental representations dynamically from real-time vehicle sensor data. Unlike static \ac{hd} maps, these methods construct maps on the fly, reducing the dependency on pre-mapped information.

Several approaches have been developed for online \ac{hd} map construction. MapTR \cite{MapTR} and StreamMapNet \cite{Streammapnet} process full surround-view camera imagery to extract road and lane structures. MapTRv2 \cite{MapTRv2} and HDMapNet \cite{li2021hdmapnet} enhance mapping capabilities by additionally integrating LiDAR data.

These algorithms provide an alternative to traditional \ac{hd} maps by leveraging vehicle perception sensors to reconstruct essential environmental elements such as lane boundaries, pedestrian crosswalks, and drivable areas. Therefore, they are particularly well-suited to potentially benefit from additional environmental information provided by our AID4AD dataset.

We propose the approach depicted in \Cref{fig:algofigdataflow} upper section, where we combine the \ac{bev} features generated from the vehicle sensor data with \ac{bev} features from our encoded AID4AD images.

\subsection{Motion Prediction}

Motion prediction is a crucial component of \ac{av} perception, as it enables the forecasting of road user trajectories based on environmental cues and their historical movement. \ac{hd} maps play a central role in these models, providing structured road network representations in either vectorized or image-based formats \cite{seff2023motionlm, wayformer, multipath, multipathpp, zhou2022hivt}. 

However, a significant drawback of \ac{hd} maps is their static nature, making them costly to update and poorly suited for rapid changes in the road environment. Accidents, roadworks, and dynamic traffic regulations may not be reflected in pre-constructed \ac{hd} maps, leading to errors in prediction models \cite{hdmapchallenges}.

Some works have attempted to replace predefined \ac{hd} maps with online-generated environmental representations to address this limitation.
\cite{mapbev1} and \cite{mapbev2} explored replacing \ac{hd} maps with online map construction algorithms, generating map features for motion prediction.
\cite{mapbev2} applied MapTR \cite{MapTR}, MapTRv2 \cite{MapTRv2}, and StreamMapNet \cite{Streammapnet} to extract \ac{bev} feature embeddings and use them in the motion prediction models HiVT \cite{zhou2022hivt} and DenseTNT \cite{densetnt}.

Our approach will follow the works of \cite{mapbev2} to integrate \ac{bev} features generated in the online map construction task into the motion prediction.

\subsection{Data Leakage}
\label{subsec:data_leak}

A critical challenge in training and evaluating map construction algorithms on nuScenes and Argoverse \cite{chang2019argoverse} is data leakage, which arises from overlapping locations between the training and evaluation sets. StreamMapNet \cite{Streammapnet} identified that the original nuScenes split—designed primarily for object detection and motion prediction—contains an 84\% overlap between training and evaluation locations. Similarly, they reported a 54\% overlap in Argoverse, raising concerns about generalization and potential overfitting to specific geographic areas. When training and evaluation regions are strictly separated, they observed an approximate 50\% drop in performance for various map construction algorithms, highlighting the extent to which models rely on memorized spatial patterns rather than generalizable learning.

To ensure a meaningful evaluation, we adopt the dataset split proposed by \cite{Roddick_2020_CVPR}, which minimizes location redundancy between training and evaluation sets. This adjusted split provides a more reliable assessment of aerial imagery-based models and prevents inflated performance estimates due to unintentional geographic bias.
\section{AID4AD Dataset Creation} 
\label{sec:AID4AD_collection}

The overall methodology is divided into two major parts: the creation of the AID4AD dataset with precisely aligned aerial imagery, and its subsequent application in \ac{av} perception tasks. This chapter focuses on the first part — the construction of the dataset and the generation of high-resolution aerial image crops that are spatially aligned with the local coordinate system of the nuScenes dataset.

To achieve this, an offset grid is computed for each recorded area, capturing the spatial discrepancies between the nuScenes base maps, which are derived from SLAM-based LiDAR reconstructions, and aerial imagery. These offset grids enable pixel-accurate alignment of the aerial images to the distorted local frame of reference used in nuScenes. Based on this alignment, \ac{bev} aerial image crops are generated for individual frames in the dataset.

The subsequent \Cref{sec:meth_algos_eval} details the application of the dataset in downstream perception tasks. These include an evaluation of map construction performance using aligned versus unaligned imagery, and a motion prediction task in which AID4AD serves as a substitute for HD maps.

Before introducing the alignment procedure itself, we first motivate the need for a dedicated correction workflow to precisely match aerial images to the local nuScenes coordinate system.

\begin{figure*}[!t]
    \centering 
    \includegraphics[width=\textwidth]{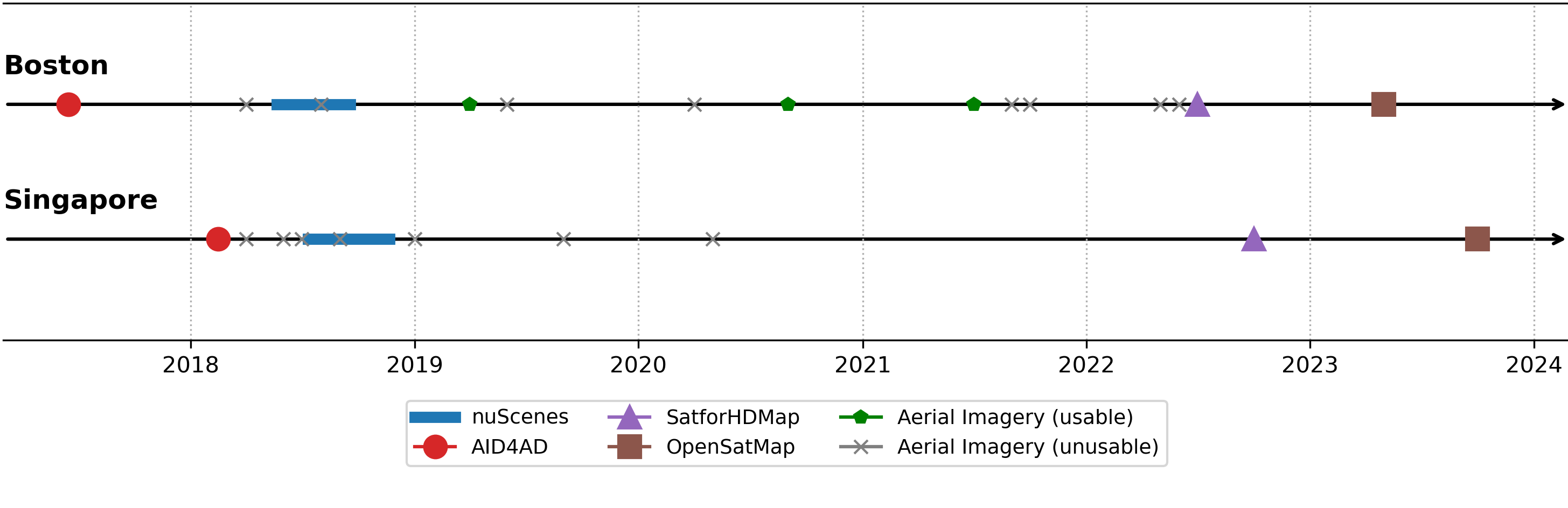} 
    \caption{Temporal alignment of nuScenes recordings with aerial imagery dates from AID4AD, SatforHDMap, and OpenSatMap. AID4AD uses imagery closely matching the original recording periods, supporting realistic research in automated driving.} 
    \label{fig:timelines} 
\end{figure*}

\subsection{Discrepancy Between Dataset Maps and Aerial Imagery}
\label{subsec:discrep}

\begin{figure}
    \centering 
    \includegraphics[width=0.48\textwidth]{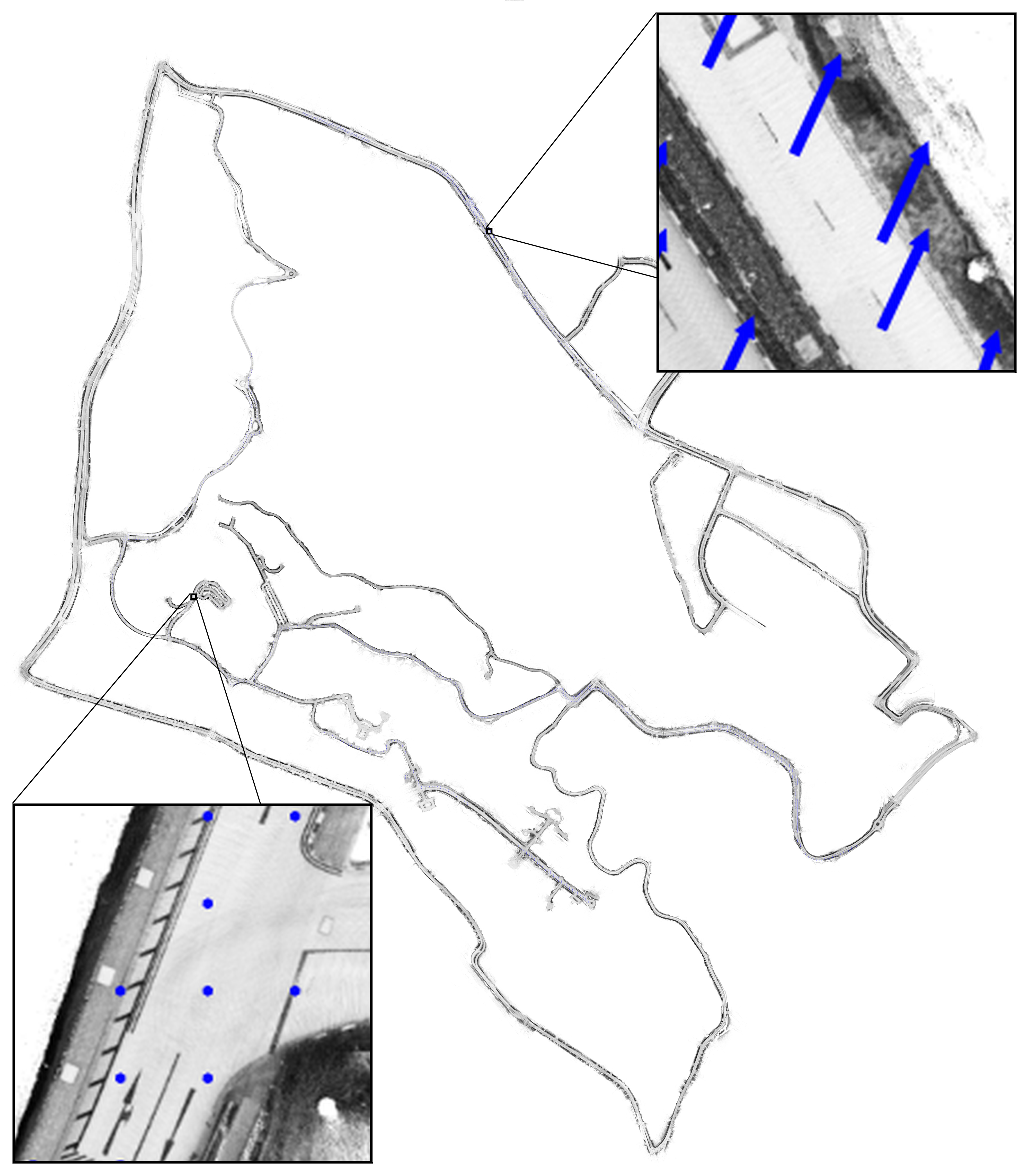} \caption{Illustration of positional discrepancies in the nuScenes base map of Queenstown compared with a corresponding aerial image. The arrow origins indicate local coordinates in the base map, while the arrowheads mark the center coordinates of aerial images aligned via visible road markings. Each arrow depicts the distance error in a 5m resolution grid.} \label{fig:quiver_plot} 
\end{figure}

A dedicated alignment process is necessary to achieve precise spatial correspondence between the nuScenes dataset and aerial imagery. Several factors contribute to the observed discrepancies. First, the GNSS coordinates in nuScenes (which can be obtained through the nuScenes devkit~\cite{nuscenes}) are not derived from raw GNSS data. Instead, they rely on the base maps provided within the dataset because pure GNSS-based localization often suffers from inaccuracies due to signal obstruction and multipath effects in dense urban environments~\cite{gnssinacc}.

These base maps are two-dimensional top-down representations created by aggregating LiDAR scans. They are generated via SLAM-based localization algorithms that combine LiDAR, GNSS, and IMU data to produce both the map and the ego vehicle coordinates in nuScenes. Due to GNSS localization issues in urban canyons, large-scale mapping artifacts and distortions appear in the base maps. Given that the largest base map spans roughly 3229 by 3687 meters, even minor local distortions can cause significant offsets when the map is projected. The map origin and the local coordinate system for each frame ultimately determine the GNSS coordinates exported by the devkit. Thus, GPS coordinates exported via the devkit inherit the accumulated SLAM distortions from the base map’s local coordinate system. An example of the positional discrepancies in these basemaps is shown in \Cref{fig:quiver_plot}.

On the aerial imagery side, data collected via Google Earth Engine \cite{googleearthengine} may contain additional distortions from perspective effects, owing to the altitude of the recording platform and the large coverage area in each tile. As a result, there is a spatial, non-linear mismatch between the nuScenes base maps and the aerial images.

Given these combined distortions, a specialized alignment workflow is necessary to achieve the level of precision needed to evaluate the potential of aerial images for \ac{av} perception tasks. 

\subsection{Temporal Selection of Aerial Imagery}
\label{subsec:temp_sel}

To ensure maximum consistency between the aerial imagery and the nuScenes dataset, we intentionally selected aerial images captured as close as possible to the respective recording periods of the nuScenes data in Boston and Singapore. This temporal proximity minimizes inconsistencies such as road layout changes, new construction, or seasonal effects, which can negatively impact perception tasks in automated driving.

For the Boston scenes, we selected imagery from June 2017, which is temporally close to the start of the nuScenes recordings in early 2018. For the Singapore regions, we used imagery from February 2018, which similarly aligns with the local nuScenes sequences recorded in 2018. This design decision provides researchers with data that more accurately reflects the scene geometry and appearance at the time of vehicle data collection.

In contrast, other datasets such as SatforHDMap and OpenSatMap use more recent aerial imagery captured several years after the nuScenes dataset, without temporal matching. This can introduce geometric or semantic shifts in the scene, making aligned image–scene representations less realistic and potentially misleading in downstream applications.

To illustrate the relevance of temporal alignment, \Cref{fig:timelines} provides an overview of aerial imagery availability over time for both cities. Besides the selected datasets, it also includes other aerial recordings that were considered during data collection. These additional entries are categorized into usable and unusable imagery. Unusable captures typically suffer from inadequate resolution, cloud coverage, or motion blur.
\begin{figure*}
    \centering 
    \includegraphics[width=\textwidth]{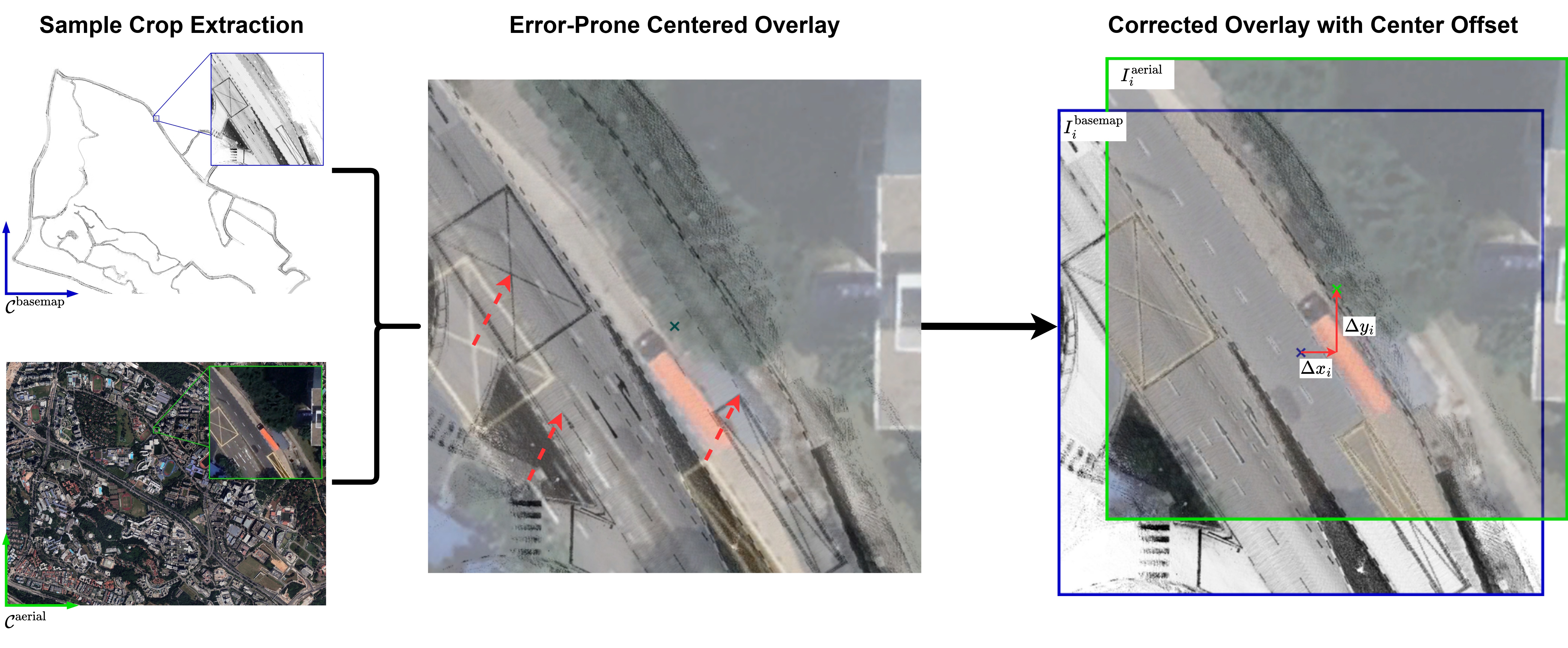} 
    \caption{Workflow for generating vehicle-centered frame crops from the base map and aerial image (left). Overlaying the crops reveals offsets (indicated as red arrows) identified through visual map priors (middle). After alignment, the horizontal offset \(\Delta x\) and the vertical offset \(\Delta y\) are visible, with the \(I^{\text{basemap}}_i\) center at the origin and \(I^{\text{aerial}}_i\) center at the tip of the arrow (right). Aerial imagery: Google Earth, Image © 2024 Maxar Technologies}
    \label{fig:overview_workflow} 
\end{figure*}
\subsection{Satellite Image--Point Cloud Map Registration} \label{subsec:meth_img_reg}

This section introduces the proposed workflow to align aerial imagery with the local coordinate system of the nuScenes dataset. The approach combines a structured coordinate representation, automated image registration, and manual quality validation to produce high-precision alignment results suitable for \ac{av} perception research.

\textbf{Coordinate Systems and Input Preparation}

For each area of interest, we utilize its basemap and aerial image, both of which are represented as 2D raster images. 
Their respective pixel coordinate systems can be transformed into metric coordinate systems \( \mathcal{C}^{\text{basemap}} \), which is the local coordinate system of the nuScenes dataset and \( \mathcal{C}^{\text{aerial}} \) using a factor of 0.15\,m/px. 
Since both images share a common geographic origin and were extracted using matching spatial extents, a coarse alignment of the two metric systems can be assumed. 
However, distortions inherent to both sources introduce spatial deviations, as detailed in Section~\ref{subsec:discrep}.

To estimate these deviations, we uniformly sample \( N \) frames across the nuScenes dataset such that the driven regions are spatially well-covered. 
For each frame \( i \in \{1, \ldots, N\} \), we extract the image crops \(I^{\text{basemap}}_i\) and \(I^{\text{aerial}}_i\) centered on the ego-vehicle’s position \( (x_i, y_i) \in \mathcal{C}^{\text{basemap}} \), which also defines the crop center in \( \mathcal{C}^{\text{aerial}} \).

\textbf{Alignment Error Identification via Image Matching}

Rather than correcting the alignment, our objective is to estimate the local pixel-wise misalignment between the map and aerial image domains. 
To do this, we iteratively compute the offset \((\Delta x_i, \Delta y_i)\) that maximizes the mutual information score (MI) \cite{mutualinformation} between the image crops of each frame \(i\):
\begin{equation}
(\Delta x_i, \Delta y_i) = \arg\max_{\Delta x, \Delta y} \ \operatorname{MI}(I^{\text{basemap}}_i, I^{\text{aerial}}_i(\Delta x, \Delta y))
\end{equation}

where \( I^{\text{aerial}}_i(\Delta x, \Delta y) \) denotes the aerial image crop shifted by the horizontal shift \(\Delta x\) and vertical shift \(\Delta y\) in pixel space.

To improve computational efficiency and suppress spurious matches, the alignment is constrained to a shift window defined by \( \Delta x, \Delta y \in [-s_{\max}, s_{\max}] \), where \( s_{\max} \) denotes the maximum allowed shift.

To enhance structural consistency, both image crops are preprocessed using Canny edge detection \cite{canny1986computational} and Contrast Limited Adaptive Histogram Equalization (CLAHE) \cite{zuiderveld1994contrast}. 

\textbf{Manual Quality Inspection and Labeling}

Although mutual information provides robust results in most cases, repetitive structures such as similar lane markings may lead to ambiguous alignments. 
To ensure high-quality supervision for future research, each automatically estimated alignment is manually reviewed.
For every sample, a visual overlay of the aerial crop on the base map is generated. 

A human annotator inspects the overlay and assigns a binary label indicating whether the alignment is correct or misaligned. 
Only validated samples are retained in the final dataset. 

The resulting set of validated alignments will be published with the dataset and serves as a ground-truth benchmark for the development and evaluation of future alignment methods.

\textbf{Vehicle Coordinates to Offset Mapping}

The validated frame-level offsets are aggregated into spatial grids for the \(x\)- and \(y\)-axis. Each validated offset is inserted into the corresponding 5\,m grid cell based on the ego vehicle's position in metric space. If multiple offsets fall into the same cell, their values are averaged. This process yields a sparse grid, as not every cell is covered by a validated frame. To obtain dense offset fields, the sparse grids are then interpolated to fill missing values.

This enables mapping any ego-vehicle coordinate within the dataset to its corresponding local misalignment in the aerial image. 
An example offset grid is shown in Figure~\ref{fig:quiver_plot}, where each arrow visualizes a displacement vector \((\Delta x_i, \Delta y_i)\) derived from the alignment process.

\subsection{Deviation Evaluation:} 
To demonstrate the alignment precision achieved by our AID4AD dataset and the underlying workflow, we conduct a comparative evaluation against two existing aerial datasets. Specifically, we assess how well the aerial images from each dataset align spatially with the nuScenes coordinate system. We randomly sample a fixed number of frames from each nuScenes region and retrieve the corresponding aerial image from AID4AD (our proposed dataset), SatforHDMap, and OpenSatMap.

At the time of writing, no official tool was available for generating frame-aligned aerial imagery from OpenSatMap. However, the authors recommend extracting GNSS coordinates for each frame using the nuScenes devkit. Following this guidance, we center aerial image crops on the error-prone GNSS positions extracted via the devkit (see \Cref{subsec:discrep}). No further alignment is applied, allowing us to evaluate the raw positioning accuracy of the original GNSS data. Future versions of OpenSatMap may introduce post-processing tools to improve alignment.

SatforHDMap provides aerial images that are cropped per frame based on GNSS-estimated positions and subsequently refined through a coarse linear alignment step. These images are used in their available form to evaluate the effectiveness of SatforHDMap’s lightweight correction approach in comparison to our proposed method. The impact of residual alignment errors is analyzed in the quantitative results section.

For AID4AD, we use aerial images that have been aligned through our dedicated workflow with respect to the nuScenes local coordinate system, aiming for high alignment precision.

Each aerial image is manually overlaid with the corresponding 2D  pointcloud map area, center-cropped around the ego vehicle. This 2D map is provided with the nuScenes dataset and reflects the static environment structure in the local coordinate system. By carefully matching persistent scene structures (e.g., intersections, road surfaces, building outlines) between the aerial and \ac{bev} images, we manually determine the translational offset in meters required to align the two views (compare to \Cref{fig:overview_workflow}). These per-frame offsets are recorded and used to compute both average and maximum distance errors for each dataset.

This quantitative analysis enables a direct comparison of alignment accuracy across datasets and is designed to assess whether the AID4AD alignment workflow achieves higher precision than the coarse correction methods applied in prior works. The results of this comparison are discussed in \Cref{subsec:SatImgRegRes}.

\begin{figure*}
    \centering 
    \includegraphics[width=\textwidth]{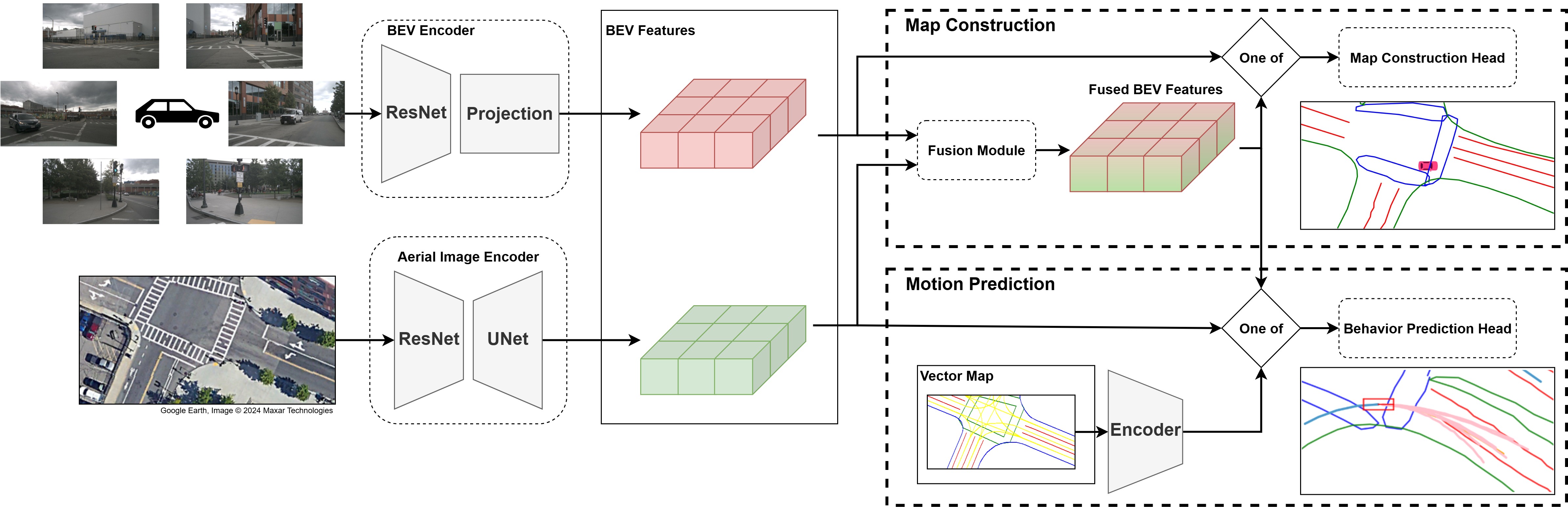} 
    \caption{Overview of the dataflows used in the evaluation. The top branch illustrates the extraction of \ac{bev} features from camera images, whereas the bottom branch depicts the encoding of aerial images. These features can then be fused for map construction or used separately for motion prediction.} 
    \label{fig:algofigdataflow} 
\end{figure*}

\section{Application of AID4AD on Perception Tasks} 
\label{sec:meth_algos_eval}

Having established our dataset and alignment workflow in \Cref{sec:AID4AD_collection}, we now employ the resulting high-resolution aerial images to assess their utility in two core tasks of \ac{av} perception: map construction and motion prediction. Specifically, we employ a ResNet-based backbone \cite{resnet} combined with a U-Net \cite{unet} decoding stage to extract multi-scale feature representations from the aerial images. Although our primary evaluation is performed on a revised version of the nuScenes split (as in \cite{Roddick_2020_CVPR}), we also provide results on the original nuScenes split to facilitate a direct comparison with prior works. This allows us to quantify the impact of precisely aligned aerial images under the proposed split and in the context of prior benchmark results.

\subsection{Application in Online Map Construction} 
\label{subsec:meth_map_constr}
To demonstrate the practical advantages of the AID4AD dataset, we begin by focusing on map construction. Specifically, we integrate the new data into the SatforHDMap codebase, adapting the code to operate on the revised nuScenes split proposed in \cite{Roddick_2020_CVPR}, mitigating potential data leakage. Simultaneously, we also evaluate SatforHDMap on the original nuScenes split to provide a comparison for previous works. AID4AD is provided in a format that directly replaces the SatforHDMap data.

We further assess the performance of AID4AD in conjunction with StreamMapNet~\cite{Streammapnet}, a more recent map construction algorithm. 
The upper section of \Cref{fig:algofigdataflow} visualizes the exemplary modifications to the data flow for both algorithms. In SatforHDMap, the authors' proposed Satellite Fusion Module is used as it is. For the integration of our data in StreamMapNet, we decided on a simple approach of stacking the \ac{bev} features generated based on the vehicle sensors with the \ac{bev} features encoded from our AID4AD images and performing a convolutional fusion.

Following previous studies, we evaluate the SatforHDMap algorithm using \ac{iou} scores and StreamMapNet based on \ac{map}, as proposed in \cite{li2021hdmapnet}, calculated under the thresholds of \textit{\{0.5m, 1.0m, and 1.5m\}}. 

\subsection{Application in Motion Prediction}
\label{subsec:meth_traj_pred}
As a second application, we investigate the effectiveness of AID4AD in the context of motion prediction using the HiVT trajectory forecasting model~\cite{zhou2022hivt}. The baseline configuration of HiVT, illustrated in the bottom-right section of \Cref{fig:algofigdataflow}, relies on structured vectorized HD maps to encode environmental context. We compare this standard input modality against two AID4AD-based alternatives inspired by the approach in~\cite{mapbev2}.

To systematically assess the influence of different environment representations, we train HiVT under four input configurations. In the baseline setting, the model utilizes the original vectorized \ac{hd} maps from nuScenes as structured environmental input. To isolate the effect of such context, we further train the model without any map-based input. In the third configuration, the HD map is replaced with \ac{bev} features derived from aerial imagery using our AID4AD-based online map construction pipeline. Finally, we incorporate a fused scene representation that combines aerial imagery with ego-vehicle surround-view observations.

In line with prior studies, we assess the performance of HiVT using commonly adopted metrics for motion prediction, such as \ac{minade}, \ac{minfde}, and \ac{mr}.
\section{Results}
\label{sec:results}

\subsection{Satellite Image--Point Cloud Map Registration Workflow}
\label{subsec:SatImgRegRes}

The following section presents the results of our satellite image–point cloud map registration workflow. We divide the evaluation into two parts: first, we assess the coverage and reliability of the automatic alignment process itself; second, we compare the spatial alignment accuracy of our resulting dataset against previous works.

\subsubsection{Automatic Alignment Success Rate}


To evaluate the performance of our automatic alignment workflow, we conducted a systematic visual inspection of all automatically aligned image-frame pairs. This step was necessary, as the current version of the workflow does not yet consistently achieve the precision desired for downstream perception tasks in automated driving.

To support the inspection, the alignment workflow generated visual overlays for each frame, where the aerial image crop, rendered with reduced saturation, was superimposed on the corresponding 2D point cloud map area. These overlays were reviewed by a human annotator, and frames exhibiting visible misalignment were excluded from further processing. A representative example of a well-aligned overlay is shown on the right in \Cref{fig:overview_workflow}, illustrating the expected visual consistency between the two modalities.

This filtering step resulted in a set of high-quality aerial image-frame pairs that serve two purposes: first, as a benchmark to assess the current capabilities of the automatic alignment; and second, as a ground truth dataset for further research. In particular, it enables the investigation of how precisely aligned aerial imagery can be integrated into perception algorithms for automated driving.

Our analysis showed that approximately 68.6\% of the automatically aligned frames passed the inspection, highlighting the workflow’s promising potential while also indicating clear opportunities for improvement in future iterations.

\subsubsection{Spatial Alignment Accuracy}

\begin{table}[t]
    \centering
    \caption{Comparison of Datasets Based on Aerial-to-Local Distance Error (ALDE), Recording Date, Resolution, and Annotations.}
    \renewcommand{\arraystretch}{1.2} 
    \begin{tabular}{l|cc|c|c}
        \toprule
        Dataset & \multicolumn{2}{c|}{ALDE (m)} & Resolution & Annotations \\
        & Mean & Max & (m/px) & Type \\
        \midrule
        SatforHDMap\cite{satforhd}      & 1.46 & 3.04                       & ~0.3 & - \\
        OpenSatMap\cite{opensatmap}     & 2.80 & 6.44   & \textbf{0.15} &  OpenSatMap \\
        AID4AD (Ours)                    & \textbf{0.16} & \textbf{0.57}     & \textbf{0.15} & nuScenes \\
        \bottomrule
    \end{tabular}
    \label{tab:dataset_comparison}
\end{table}

To quantify the spatial alignment precision, we introduce the \textit{Aerial-to-Local Distance Error (ALDE)}, defined as the Euclidean norm of the translational offsets \( (\Delta x_i, \Delta y_i) \) required to align the aerial image crop to the corresponding base map crop for each of the \( N \) evaluated frames. We report both the mean and maximum values:

\begin{equation}
\text{ALDE}_{\text{mean}} = \frac{1}{N} \sum_{i=1}^{N} \sqrt{(\Delta x_i)^2 + (\Delta y_i)^2}
\end{equation}
\begin{equation}
\text{ALDE}_{\text{max}} = \max_{i} \sqrt{(\Delta x_i)^2 + (\Delta y_i)^2}
\end{equation}

A comparison with previous aerial datasets is shown in \Cref{tab:dataset_comparison}. AID4AD achieves significantly lower ALDE values, outperforming prior methods by a factor of five to ten in both mean and maximum.

OpenSatMap and AID4AD are equivalent regarding resolution measured in meters per pixel. However, the per-frame crops from SatforHDMap data restrict its maximum resolution to only 0.3 meters per pixel compared to 0.15 for AID4AD and OpenSatMap.

A key distinction among these datasets is the availability of annotations. The OpenSatMap project dedicated substantial financial and time resources to annotating its aerial images, resulting in the most comprehensive road extraction dataset available. In addition to annotations related to road structure, each aerial image is classified according to its quality, which is influenced by factors such as occlusion from nearby buildings, trees, or perspective.

The exact alignment allows AID4AD to effectively leverage the annotations from nuScenes, as illustrated in \Cref{fig:alignment}.

\subsection{AID4AD in Map Construction}
\label{subsec:map_constr_eval}

\begin{table}[t]
    \centering
    \caption{Comparison of IoU Scores for HDMapNet and SatforHDMap with Original Input, and SatforHDMap with Input Replaced by OpenSatMap or AID4AD, Evaluated on Divider, Crossing, and Boundary Classes}
    \begin{tabular}{l|ccc|c}
        \toprule
        \multirow{2}{*}{Algorithm} & \multicolumn{3}{c|}{IoU Score (\%)} & \multirow{2}{*}{All} \\
        & Divider & Crossing & Boundary &  \\
        \midrule
        \multicolumn{5}{c}{\textit{original split}} \\
        \midrule
        HDMapNet & 40.6 & 18.7 & 39.5 & 32.9 \\
        SatforHDMap & 54.9 & 53.4 & 52.9 & 53.7 \\[0.1em]   
        SatforHDMap & \multirow{2}{*}{50.2} & \multirow{2}{*}{53.2} & \multirow{2}{*}{49.4} & \multirow{2}{*}{50.9} \\
        + OpenSatMap & & & \\[0.1em]   
        SatforHDMap & \multirow{2}{*}{\textbf{60.2}} & \multirow{2}{*}{\textbf{66.5}} & \multirow{2}{*}{\textbf{60.4}} & \multirow{2}{*}{\textbf{62.4}} \\
        + AID4AD & & & \\
        \midrule
        \midrule
        \multicolumn{5}{c}{\textit{new split}} \\
        \midrule
        SatforHDMap & 14.9 & 3.8 & 13.1 & 10.6 \\[0.1em]   
        SatforHDMap & \multirow{2}{*}{\textbf{29.0}} & \multirow{2}{*}{\textbf{38.8}} & \multirow{2}{*}{\textbf{32.0}} & \multirow{2}{*}{\textbf{33.3}} \\
        + AID4AD & & & \\
        \midrule
        $\Delta mIoU$ & +14.1 & +35.0 & +18.9 & +22.7 \\        
        \bottomrule
    \end{tabular}
    \label{tab:iou_comparison}
\end{table}

To analyze AID4AD's impact on map construction, we evaluate its performance in two key comparisons. First, we assess the effect of precise aerial image alignment versus coarse alignment using the same algorithm and fusion module as in SatforHDMap's work. Second, we integrate AID4AD into StreamMapNet using a simpler convolutional fusion approach.

In \Cref{tab:iou_comparison}, we present the results of the SatforHDMap algorithm when combined with our AID4AD dataset, comparing it with HDMapNet, SatforHDMap using their works data, and OpenSatMap images for the old nuScenes split. We also present the results for the new split, comparing the SatforHDMap algorithm using the SatforHD dataset and our AID4AD dataset. The results show a significant increase of, on average, +22.7\% IoU score when using our aerial images compared to the coarsely aligned and lower resolution images from SatforHDMap. 

We further evaluate the impact of the AID4AD data on the map construction algorithm StreamMapNet. Our results demonstrate a significant performance improvement of +14.7\% in the \ac{map} when the algorithm is supplemented with our AID4AD images.
The output from the AID4AD encoder and the fusion module, in the form of \ac{bev} features, is saved for later use in the motion prediction task.

\begin{table}[t]
    \centering
    \caption{Performance of the StreamMapNet \cite{Streammapnet} Online Map Construction Algorithm in the Original Setup and with AID4AD Image Fusion}
    \begin{tabular}{l|ccc|c}
        \toprule
        \multirow{2}{*}{Algorithm} & \multicolumn{3}{c|}{Average Precision (\%)} & \multirow{2}{*}{\ac{map}} \\
        & Divider & Crossing & Boundary &  \\
        \midrule
        StreamMapNet & 29.3 & 32.2 & 40.8 & 34.1 \\[0.1em]   
        StreamMapNet  & \multirow{2}{*}{\textbf{32.9}} & \multirow{2}{*}{\textbf{59.9}} & \multirow{2}{*}{\textbf{53.6}} & \multirow{2}{*}{\textbf{48.8}} \\
        + AID4AD & & & \\
        \midrule
        $\Delta mAP$ & +3.6 & +27.7 & +12.8 & +14.7 \\
        \bottomrule
    \end{tabular}
    \label{tab:streammapnet_res}
\end{table}

\subsection{AID4AD in Motion Prediction}
\label{subsec:traj_res}

Finally, we evaluated the effect of AID4AD on the motion prediction task as described in \Cref{subsec:meth_traj_pred}.

Table \ref{tab:prediction_metrics} summarizes the prediction performance under varying environmental context sources. Using features derived solely from AID4AD imagery narrows the performance gap to \ac{hd} maps substantially, with an average degradation of only 3\% compared to 12\% when no environmental context is used. When combining AID4AD with surround-view-based map construction via StreamMapNet, the system even surpasses the \ac{hd} map baseline, achieving a 2\% average improvement across all metrics. These results indicate that aerial imagery, especially when fused with onboard perception, offers a scalable and competitive alternative to \ac{hd} maps in motion prediction tasks.

\begin{table}
    \centering
    \caption{Comparison of HiVT Motion Prediction Performance Using HD Maps, Stored \ac{bev} Features, or No Environmental Input}
    \begin{tabular}{l|lll}
        \toprule
        Environmental & \multirow{2}{*}{\ac{minade} ↓} & \multirow{2}{*}{\ac{minfde} ↓} & \multirow{2}{*}{\ac{mr} ↓} \\
        Information Source & & & \\
        \midrule
        \ac{hd} Maps & \textbf{0.385} & 0.800 & 0.088 \\[0.1em]
        None & 0.411\,\textcolor{DarkRed}{(+7\%)} & 0.880\,\textcolor{DarkRed}{(+10\%)} & 0.104\,\textcolor{DarkRed}{(+18\%)} \\[0.1em]
        AID4AD Features & 0.388\,\textcolor{DarkRed}{(+1\%)} & 0.817\,\textcolor{DarkRed}{(+2\%)} & 0.094\,\textcolor{DarkRed}{(+7\%)} \\[0.1em]        
        StreamMapNet & \multirow{2}{*}{0.386\,\textcolor{DarkGray}{(0\%)}} & \multirow{2}{*}{\textbf{0.790}\,\textcolor{DarkGreen}{(-1\%)}} & \multirow{2}{*}{\textbf{0.084}\,\textcolor{DarkGreen}{(-5\%)}} \\
        + AID4AD Features & & & \\
        \bottomrule        
    \end{tabular}
    \label{tab:prediction_metrics}
\end{table}
\section{Conclusion}
\label{sec:conclusion}
This study establishes a foundation for integrating aerial imagery into the perception stack of \acp{av}. We presented AID4AD, a dataset of spatially aligned aerial images designed to complement the nuScenes dataset, enabling systematic research on the use of \ac{bev} imagery in \ac{av} systems. Central to this contribution is our proposed alignment workflow, which utilizes SLAM-based point cloud maps to register aerial images in the vehicle’s local coordinate system, thereby correcting for localization and projection errors.

To ensure spatial precision, the workflow includes a manual quality control step, resulting in a high-quality subset of alignment pairs. This ground-truth set is released to support further research on automated alignment methods and aerial image integration.

We also proposed and evaluated an integration strategy for using aerial imagery in \ac{av} perception algorithms, demonstrating its effectiveness in two representative tasks. In online map construction, aerial imagery complements sensor data, improving the generation of structured map representations by 15–23\%. In trajectory prediction, it replaces \ac{hd} maps as the environmental context, yielding a 2\% performance gain. These results confirm that aerial imagery can serve as a structured, scalable, and updatable source of environmental knowledge in scenarios where \ac{hd} maps are unavailable or impractical.

Looking forward, a fully automatic, high-precision alignment workflow could enable georeferencing of local maps and facilitate real-time updates via drone-collected imagery. This opens new research directions in dynamic map generation, scalable localization, and environment-aware \ac{av} perception.

AID4AD establishes a foundation for leveraging semantic cues from aerial imagery in downstream perception tasks. The global perspective offered by aerial views can complement vehicle-centric observations in applications such as drivable area estimation, scene understanding, or context-aware planning.

To foster continued research, we publicly release AID4AD alongside our alignment tools, evaluation framework, and pretrained models, laying the groundwork for aerial imagery as a core enabler of future \ac{av} perception systems.

\ifCLASSOPTIONcaptionsoff
  \newpage
\fi

\bibliographystyle{IEEEtran}
\bibliography{main}

\begin{thebibliography}{10}
\providecommand{\url}[1]{#1}
\csname url@samestyle\endcsname
\providecommand{\newblock}{\relax}
\providecommand{\bibinfo}[2]{#2}
\providecommand{\BIBentrySTDinterwordspacing}{\spaceskip=0pt\relax}
\providecommand{\BIBentryALTinterwordstretchfactor}{4}
\providecommand{\BIBentryALTinterwordspacing}{\spaceskip=\fontdimen2\font plus
\BIBentryALTinterwordstretchfactor\fontdimen3\font minus
  \fontdimen4\font\relax}
\providecommand{\BIBforeignlanguage}[2]{{%
\expandafter\ifx\csname l@#1\endcsname\relax
\typeout{** WARNING: IEEEtran.bst: No hyphenation pattern has been}%
\typeout{** loaded for the language `#1'. Using the pattern for}%
\typeout{** the default language instead.}%
\else
\language=\csname l@#1\endcsname
\fi
#2}}
\providecommand{\BIBdecl}{\relax}
\BIBdecl

\bibitem{WorldView3_ESA}
{European Space Agency (ESA)}, ``Worldview-3 mission - earth observation
  gateway,'' \url{https://earth.esa.int/eogateway/missions/worldview-3}, 2024,
  accessed July 2025.

\bibitem{Jasnoch_2023}
T.~Jasnoch and T.~Scholz, ``Accuracy assessment of rtk/ppk uav-photogrammetry
  projects,'' \emph{Journal of Applied Remote Sensing}, vol.~17, no.~03, p.
  035001, 2023.

\bibitem{PrecisionAg_Review}
C.~Zhang and J.~M. Kovacs, ``The application of small unmanned aerial systems
  for precision agriculture: a review,'' \emph{Precision Agriculture}, vol.~13,
  pp. 693--712, 2012.

\bibitem{RemoteSensing_Applications}
C.~J. Tucker and J.~R.~G. Townshend, ``Strategies for monitoring tropical
  deforestation using satellite data,'' \emph{International Journal of Remote
  Sensing}, vol.~21, no. 6-7, pp. 1461--1472, 2000.

\bibitem{kutsch_tumdotmuc_2024}
A.~Kutsch, M.~Margreiter, and K.~Bogenberger, ``Tumdot–muc: Data collection
  and processing of multimodal trajectories collected by aerial drones,''
  \emph{Data Science for Transportation}, vol.~6, no.~2, 2024.

\bibitem{DOP20_BKG}
{Federal Agency for Cartography and Geodesy (BKG)}, ``Digitale orthophotos
  bodenauflösung 20 cm (dop20),''
  \url{https://gdz.bkg.bund.de/index.php/default/digitale-orthophotos-bodenauflosung-20-cm-dop20.html},
  2023, accessed July 2025.

\bibitem{OS_UK}
{Ordnance Survey}, ``Os mastermap imagery layer,''
  \url{https://www.ordnancesurvey.co.uk/products/os-mastermap-imagery-layer},
  2023, accessed July 2025.

\bibitem{BDORTHO_FR}
{Institut Géographique National (IGN)}, ``Bd ortho® – aerial imagery of
  metropolitan france,'' \url{https://geoservices.ign.fr/bdortho}, 2023,
  accessed July 2025.

\bibitem{multipath}
Y.~Chai, B.~Sapp, M.~Bansal, and D.~Anguelov, ``Multipath: Multiple
  probabilistic anchor trajectory hypotheses for behavior prediction,'' in
  \emph{Proceedings of the Conference on Robot Learning}, ser. Proceedings of
  Machine Learning Research, L.~P. Kaelbling, D.~Kragic, and K.~Sugiura, Eds.,
  vol. 100.\hskip 1em plus 0.5em minus 0.4em\relax PMLR, 30 Oct--01 Nov 2020,
  pp. 86--99.

\bibitem{multipathpp}
B.~Varadarajan, A.~Hefny, A.~Srivastava, K.~S. Refaat, N.~Nayakanti, A.~Cornman
  \emph{et~al.}, ``Multipath++: Efficient information fusion and trajectory
  aggregation for behavior prediction,'' in \emph{2022 International Conference
  on Robotics and Automation (ICRA)}, 2022, pp. 7814--7821.

\bibitem{zhou2022hivt}
Z.~Zhou, L.~Ye, J.~Wang, K.~Wu, and K.~Lu, ``Hivt: Hierarchical vector
  transformer for multi-agent motion prediction,'' in \emph{Proceedings of the
  IEEE/CVF Conference on Computer Vision and Pattern Recognition (CVPR)}, 2022.

\bibitem{hdmapchallenges}
G.~Elghazaly, R.~Frank, S.~Harvey, and S.~Safko, ``High-definition maps:
  Comprehensive survey, challenges, and future perspectives,'' \emph{IEEE Open
  Journal of Intelligent Transportation Systems}, vol.~4, pp. 527--550, 2023.

\bibitem{nuscenes}
H.~Caesar, V.~Bankiti, A.~H. Lang, S.~Vora, V.~E. Liong, Q.~Xu \emph{et~al.},
  ``nuscenes: A multimodal dataset for autonomous driving,'' in \emph{CVPR},
  2020.

\bibitem{googleearthengine}
N.~Gorelick, M.~Hancher, M.~Dixon, S.~Ilyushchenko, D.~Thau, and R.~Moore,
  ``Google earth engine: Planetary-scale geospatial analysis for everyone,''
  \emph{Remote Sensing of Environment}, 2017.

\bibitem{aid_map_constr}
S.~He and H.~Balakrishnan, ``Lane-level street map extraction from aerial
  imagery,'' in \emph{Proceedings of the IEEE/CVF Winter Conference on
  Applications of Computer Vision (WACV)}, January 2022, pp. 2080--2089.

\bibitem{aid_map_constr2}
J.~Yao, X.~Pan, T.~Wu, and X.~Zhang, ``Building lane-level maps from aerial
  images,'' in \emph{ICASSP 2024-2024 IEEE International Conference on
  Acoustics, Speech and Signal Processing (ICASSP)}.\hskip 1em plus 0.5em minus
  0.4em\relax IEEE, 2024, pp. 3890--3894.

\bibitem{satforhd}
W.~Gao, J.~Fu, Y.~Shen, H.~Jing, S.~Chen, and N.~Zheng, ``Complementing onboard
  sensors with satellite map: A new perspective for hd map construction,'' in
  \emph{IEEE International Conference on Robotics and Automation}, 2024.

\bibitem{opensatmap}
H.~Zhao, L.~Fan, Y.~Chen, H.~Wang, yuran Yang, X.~Jin \emph{et~al.},
  ``Opensatmap: A fine-grained high-resolution satellite dataset for
  large-scale map construction,'' in \emph{Thirty-eighth Conference on Neural
  Information Processing Systems}, 2024.

\bibitem{li2021hdmapnet}
Q.~Li, Y.~Wang, Y.~Wang, and H.~Zhao, ``Hdmapnet: An online hd map construction
  and evaluation framework,'' in \emph{2022 International Conference on
  Robotics and Automation (ICRA)}, 2022, pp. 4628--4634.

\bibitem{waymo_perc}
P.~Sun, H.~Kretzschmar, X.~Dotiwalla, A.~Chouard, V.~Patnaik, P.~Tsui
  \emph{et~al.}, ``Scalability in perception for autonomous driving: Waymo open
  dataset,'' in \emph{Proceedings of the IEEE/CVF Conference on Computer Vision
  and Pattern Recognition (CVPR)}, June 2020.

\bibitem{Geiger2013IJRR}
A.~Geiger, P.~Lenz, C.~Stiller, and R.~Urtasun, ``Vision meets robotics: The
  kitti dataset,'' \emph{International Journal of Robotics Research (IJRR)},
  2013.

\bibitem{MapTR}
B.~Liao, S.~Chen, X.~Wang, T.~Cheng, Q.~Zhang, W.~Liu \emph{et~al.}, ``Maptr:
  Structured modeling and learning for online vectorized hd map construction,''
  in \emph{International Conference on Learning Representations}, 2023.

\bibitem{Streammapnet}
T.~Yuan, Y.~Liu, Y.~Wang, Y.~Wang, and H.~Zhao, ``Streammapnet: Streaming
  mapping network for vectorized online hd map construction,'' in
  \emph{Proceedings of the IEEE/CVF Winter Conference on Applications of
  Computer Vision (WACV)}, January 2024, pp. 7356--7365.

\bibitem{MapTRv2}
B.~Liao, S.~Chen, Y.~Zhang, B.~Jiang, Q.~Zhang, W.~Liu \emph{et~al.},
  ``Maptrv2: An end-to-end framework for online vectorized hd map
  construction,'' \emph{International Journal of Computer Vision}, pp. 1--23,
  2024.

\bibitem{seff2023motionlm}
A.~Seff, B.~Cera, D.~Chen, M.~Ng, A.~Zhou, N.~Nayakanti \emph{et~al.},
  ``Motionlm: Multi-agent motion forecasting as language modeling,'' 2023.

\bibitem{wayformer}
N.~Nayakanti, R.~Al-Rfou, A.~Zhou, K.~Goel, K.~S. Refaat, and B.~Sapp,
  ``Wayformer: Motion forecasting via simple \& efficient attention networks,''
  2022.

\bibitem{mapbev1}
X.~Gu, G.~Song, I.~Gilitschenski, M.~Pavone, and B.~Ivanovic, ``Producing and
  leveraging online map uncertainty in trajectory prediction,'' in
  \emph{Proceedings of the IEEE/CVF Conference on Computer Vision and Pattern
  Recognition (CVPR)}, 2024.

\bibitem{mapbev2}
------, ``Accelerating online mapping and behavior prediction via direct bev
  feature attention,'' in \emph{European Conference on Computer Vision (ECCV)},
  2024.

\bibitem{densetnt}
J.~Gu, C.~Sun, and H.~Zhao, ``Densetnt: End-to-end trajectory prediction from
  dense goal sets,'' in \emph{Proceedings of the IEEE/CVF International
  Conference on Computer Vision}, 2021, pp. 15\,303--15\,312.

\bibitem{chang2019argoverse}
M.-F. Chang, J.~Lambert, P.~Sangkloy, J.~Singh, S.~Bak, A.~Hartnett
  \emph{et~al.}, ``Argoverse: 3d tracking and forecasting with rich maps,'' in
  \emph{Proceedings of the IEEE/CVF Conference on Computer Vision and Pattern
  Recognition}, 2019, pp. 8748--8757.

\bibitem{Roddick_2020_CVPR}
T.~Roddick and R.~Cipolla, ``Predicting semantic map representations from
  images using pyramid occupancy networks,'' in \emph{IEEE/CVF Conference on
  Computer Vision and Pattern Recognition (CVPR)}, June 2020.

\bibitem{gnssinacc}
N.~Zhu, J.~Marais, D.~Bétaille, and M.~Berbineau, ``Gnss position integrity in
  urban environments: A review of literature,'' \emph{IEEE Transactions on
  Intelligent Transportation Systems}, vol.~19, no.~9, pp. 2762--2778, 2018.

\bibitem{mutualinformation}
P.~Viola and W.~Wells, ``Alignment by maximization of mutual information,'' in
  \emph{Proceedings of IEEE International Conference on Computer Vision}, 1995,
  pp. 16--23.

\bibitem{canny1986computational}
J.~Canny, ``A computational approach to edge detection,'' \emph{IEEE
  Transactions on Pattern Analysis and Machine Intelligence}, vol. PAMI-8,
  no.~6, pp. 679--698, 1986.

\bibitem{zuiderveld1994contrast}
K.~Zuiderveld, ``Contrast limited adaptive histogram equalization,'' in
  \emph{Graphics Gems IV}.\hskip 1em plus 0.5em minus 0.4em\relax Academic
  Press, 1994, pp. 474--485.

\bibitem{resnet}
K.~He, X.~Zhang, S.~Ren, and J.~Sun, ``Deep residual learning for image
  recognition,'' in \emph{2016 IEEE Conference on Computer Vision and Pattern
  Recognition (CVPR)}, 2016, pp. 770--778.

\bibitem{unet}
O.~Ronneberger, P.~Fischer, and T.~Brox, ``U-net: Convolutional networks for
  biomedical image segmentation.'' \emph{CoRR}, vol. abs/1505.04597, 2015.

\end{thebibliography}


\begin{IEEEbiography}[{\includegraphics[width=1in,height=1.25in,clip,keepaspectratio]{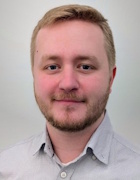}}]{Daniel Lengerer}  received the B.Eng. degree in Electrical Engineering and the M.Sc. degree in Computer Science from the Technical University of Applied Sciences Augsburg, Germany. He is currently pursuing a joint Ph.D. degree with the Chair of Traffic Engineering at the Technical University of Munich and the Augsburg University of Applied Sciences. His research focuses on perception and prediction methods for automated vehicles, with an emphasis on robust scene understanding. 
This includes investigating alternatives to high-definition maps, such as the integration of aerial imagery, and developing novel approaches to improve motion forecasting in complex environments.
\end{IEEEbiography}

\begin{IEEEbiography}[{\includegraphics[width=1in,height=1.25in,clip,keepaspectratio]{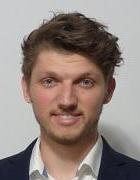}}]{Mathias Pechinger} was born in Wertingen, Bavaria, Germany, in 1993. He received the B.Eng. degree in electrical engineering and the M.Sc. degree in computer science from the Augsburg University of Applied Sciences, Germany. He completed his Ph.D. in 2023 through a joint doctoral program with the Chair of Traffic Engineering and Control, Technical University of Munich, and the Augsburg University of Applied Sciences. In the same year, he joined the Chair of Traffic Engineering and Control at the Technical University of Munich, where he leads the research group for Experimental Studies. His work focuses on hardware-in-the-loop tests for infrastructure-based guidance for automated vehicles, investigating the value added by roadside ITS-S to the overall picture of automated mobility and the impact of introducing new modes of transport to the urban environment.
\end{IEEEbiography}

\begin{IEEEbiography}[{\includegraphics[width=1in,height=1.25in,clip,keepaspectratio]{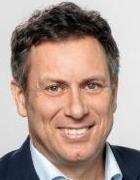}}]{Klaus Bogenberger} received the Diploma degree in civil engineering and the Ph.D. degree in traffic engineering from the Technical University of Munich, Germany, in 1996 and 2001, respectively. He was a Research Engineer with BMW Group from 2001 to 2008. He was initially responsible for traffic flow theory and models with the Department of Science and Transportation and later with the Department of Corporate Quality. From 2008 to 2011, he was the Managing Director and a Partner of the TRANSVER GmbH (Consultant Office for Transport Planning and Traffic Engineering) in Munich and Hannover. In 2012, he was appointed as a Professor of traffic engineering with the Munich University of the Federal Armed Forces. Since 2020, he has been leading the Chair of Traffic Engineering and Control, Technical University of Munich. His research interests include car-sharing systems, quality of traffic
information, and autonomous vehicles.
\end{IEEEbiography}

\begin{IEEEbiography}[{\includegraphics[width=1in,height=1.25in,clip,keepaspectratio]{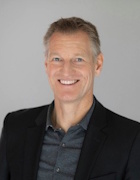}}]{Carsten Markgraf} received the degree in electrical engineering from the University of Hanover, Germany, in 1997, and the Ph.D. degree in 2002 with a thesis on “Automated driving with magnetic nails.” 
As a research assistant, he cooperated with VW to analyze infrastructure-based technologies for automating vehicle test tracks. After his Ph.D. degree, he worked in industry at ThyssenKrupp Automotive, Munich, where he developed intelligent chassis control systems. Then, he worked at ThyssenKrupp Automotive, Liechtenstein, where he developed electric power steering systems (EPS) and was responsible for developing and industrializing the electrical, electronic, and software components. 
After the first system went into mass production for the X3 BMW in 2010, he received a call from the Augsburg University of Applied Sciences to become a Professor of electrical engineering, 
where he is responsible for automatic control and dynamic systems. Today, he is Scientific Director of TTZ DataScience and Autonomous Systems at THA.
\end{IEEEbiography}

\end{document}